# Conception et usages des dispositifs d'assistance aux mouvements du membre supérieur : Revue de littérature

C. Le Goff[1,2], P. Coignard[1,4], Christine Azevedo-Coste[2], F. Geffard[5], C. Fattal[1,2,3]

1. Association Approche, Ploemeur
2. INRIA, Univ Montpellier, CNRS, Montpellier
3. USSAP Centre Bouffard-Vercelli, Perpignan
4. CMRRF Kerpape, Ploemeur
5. CEA, List, Université Paris-Saclay, Palaiseau

Auteurs correspondant : Charlotte LE GOFF
Adresse postale : Siège social CMRRF de Kerpape, BP78 - 56275 Ploemeur Cedex
Adresse e-mail : charlotte.le-goff@inria.fr
Téléphone : 02.97.82.61.74

**RÉSUMÉ** :
Cet article explore les dispositifs d'assistance aux mouvements du membre supérieur chez les personnes en situation de handicap à travers une revue systématique basée sur une méthodologie PRISMA. Les dispositifs étudiés couvrent des technologies allant des orthèses aux robots avancés, visant à compenser ou à suppléer des déficiences motrices. Les résultats mettent en lumière la diversité des usages (rééducation, activités de vie quotidienne), des segments corporels ciblés (distal, proximal, ou global), ainsi que des mécanismes et interfaces de commandes employés. Cependant, malgré cette variété de prototypes prometteurs, peu de dispositifs sont effectivement disponibles dans le commerce, limitant leur impact concret sur les utilisateurs finaux. Les technologies existantes, bien qu'efficaces pour améliorer l'autonomie fonctionnelle et la qualité de vie, présentent encore des limitations en termes d'ergonomie, de coût et de portabilité. En conclusion, cet article souligne l'importance d'une approche centrée sur l'usager et propose des pistes pour le développement de dispositifs innovants, modulaires et accessibles.

**MOTS CLÉS** : membre supérieur, technologie d'assistance, robotique, déficience motrice, activité de vie quotidienne.

**ABSTRACT** :
This article explores assistive devices for upper limb movement in individuals with disabilities through a systematic review based on the PRISMA methodology. The studied devices encompass technologies ranging from orthoses to advanced robots, aimed at compensating for or substituting motor impairments. The findings highlight the diversity of applications (rehabilitation, activities of daily living), targeted body segments (distal, proximal, or global), and employed mechanisms and control interfaces. However, despite this variety of promising prototypes, few devices are actually commercially available, limiting their tangible impact on end-users. Existing technologies, while effective in improving functional autonomy and quality of life, still face challenges related to ergonomics, cost, and portability. In conclusion, this article underscores the importance of a user-centered approach and suggests pathways for the development of innovative, modular, and accessible devices.

**INTRODUCTION :**

Dans le monde, on estime que 1,3 milliard de personnes sont en situation de handicap[1], résultant de maladies chroniques, d'accidents ou du vieillissement. La Classification Internationale du Fonctionnement, du Handicap et de la Santé (CIF)[2], met en évidence les interactions complexes entre les dimensions biologiques, sociales et environnementales, en identifiant les facteurs limitants et les ressources mobilisables pour renforcer l'autonomie et la qualité de vie. Ce cadre conceptuel est particulièrement pertinent pour comprendre les enjeux liés aux déficiences motrices, soulignant l'importance des technologies d'assistance. La déficience motrice du membre supérieur (MS) se réfère à une limitation partielle ou totale des capacités motrices des bras, avant-bras, mains ou doigts entraînant une diminution de la force, de la précision, de la coordination ou de l'amplitude des mouvements. Une prise en charge précoce, adaptée et pluridisciplinaire, incluant parfois des aides techniques, est essentielle pour limiter ces impacts. La norme ISO 9999:2022[3] définit les aides techniques comme des dispositifs conçus pour prévenir, compenser ou neutraliser une déficience. Ces aides, allant d'outils simples à des systèmes robotiques avancés, visent à améliorer l'autonomie des personnes en renforçant, assistant ou suppléant le mouvement humain, et jouent un rôle clé dans la compensation des déficiences motrices. On distingue ainsi la suppléance fonctionnelle, qui compense une perte motrice en remplaçant ou assistant directement une fonction, de la rééducation fonctionnelle qui favorise la récupération ou l'amélioration des capacités motrices par une stimulation intensive et répétée.

Cet article présente une revue de littérature sur les dispositifs d'assistance pour le MS chez les personnes en situation de handicap. Plusieurs points seront abordés sur les usages et contextes d'utilisations des dispositifs, les segments du MS et les mouvements concernés, les technologies employées, leurs commandes et interfaces de pilotage, ainsi que les pathologies adressées. Nous discuterons des avantages mais aussi des limites non résolues par ces dispositifs. Cette analyse de littérature est une base de réflexion pour contribuer au développement d'innovations technologiques au service des personnes en situation de handicap. Cet article s'appuie sur les résultats d'un travail annexe suivant une méthodologie PRISMA. Cette méthodologie exclut certains dispositifs disponibles sur le marché en France, n'ayant pas fait l'objet d'études publiées, néanmoins intéressants pour la pratique courante et cités dans l'article de C. Fattal.

**MÉTHODES :**

Les articles étudiés sont extraits des bases de données PubMed, l'Institute of Electrical and Electronics Engineers (IEEE) et PASCAL, à partir de diverses combinaisons de mots clés (tableau 1) selon une méthodologie PCC (Population, Concept, Context)[4].

| POPULATION | | | CONCEPT | CONTEXT |
|---|---|---|---|---|
| Upper limb deficiency<br>Motor impairment<br>Motor skills disorders<br>Upper limb dysfunction<br>Upper extremity injury<br>Upper extremity physiopathology<br>Arm dysfunction / Arm injuries<br>Forearm dysfunction<br>Shoulder dysfunction<br>Elbow dysfunction<br>Wrist dysfunction<br>Hand dysfunction<br>Fingers dysfunction<br><br>Poliomyelitis<br>Postpoliomyelitis syndrome<br>Peripheral neuropathy<br>Guillain-Barré syndrome<br>Brachial plexus injury | Neuromuscular disease<br>Neuromuscular disorders<br>Muscle weakness<br>Muscular disease<br>Muscular weakness<br>Musculoskeletal disease<br>Muscular dystrophy<br>Duchenne<br>Steinert<br>Myopath*<br>Myositis<br>Atrophy<br>Myotonic dystrophy<br>Spinal muscular atrophy<br>Motor neurone disease /<br>Amyotrophic lateral sclerosis<br><br>Joint disease<br>Arthropathy<br>Arthrogryposis<br>Rheumatological disease<br>Rheumatoid arthritis<br>Rotator cuff | Spinal cord injury (tetraplegia, paraplegia)<br>Cerebrovascular disorder<br>Cerebrovascular disease<br>Stroke / post-stroke<br>Brain ischemia<br>Brain injury<br>Cerebrovascular accident<br>Vasospasm intracranial<br>Vertebral artery dissection<br>Brain infarction<br>Cerebrovascular trauma<br>Intracranial arterial disease<br>Parkinson disease<br>Cerebral palsy (diplegia, quadriplegia)<br>Cerebellar syndrome<br>Hemipleg*<br>Hemipare*<br>Locked-in syndrome<br>Multiple sclerosis | Orthotic Devices<br>Orthotic<br>Self-Help Devices<br>Robotics<br>Robot*<br>Exoskelet*<br>Arm support<br>Assistive technology<br>Assistive technologies<br>Assistive device<br>Assistive devices<br>Orthosis | Substitution<br>Assistive<br>Compensation<br>Disabled person<br>Activities of daily living |

Tableau [1] : Liste des mots-clés utilisés dans la stratégie de recherche selon la méthodologie PCC (Population, Concept, Context)

La sélection des articles pour cette revue de littérature a suivi une approche séquentielle de tri, selon la méthode décrite par *Van Der Heide et al*[5]. Au total, 168 articles ont été retenus et analysés pour lecture complète. Seuls les articles originaux incluant des études cliniques sont retenus. Les articles portant sur les dispositifs statiques, de rééducation non transposable à un usage quotidien, les prothèses, et les études non axées sur le handicap ont été exclus. Pour collecter les données, un formulaire d'extraction structuré a été utilisé, incluant des champs sur les usages et contextes d'utilisations des dispositifs, les segments du MS et les mouvements impliqués, les technologies employées, leurs commandes et interfaces de pilotages, ainsi que la pathologie adressée.

**RÉSULTATS :**

L'ensemble des résultats s'appuieront sur la figure suivante :

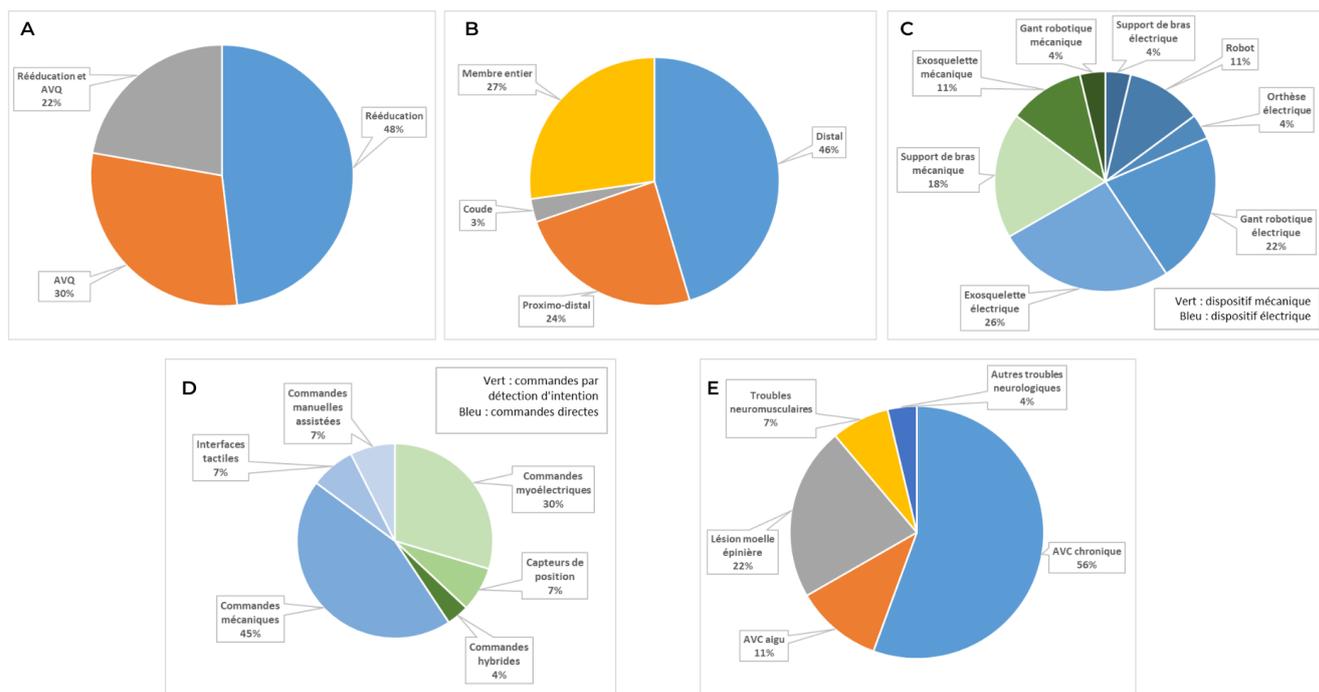

Figure [1] : Représentation graphique des données sur les dispositifs étudiés avec (A) les usages et contextes d'utilisation des dispositifs ; (B) les segments du membre supérieur et les mouvements impliqués ; (C) les technologies employées ; (D) les commandes et interfaces de pilotage ; (E) les pathologies adressées.

**Usages et contextes d'utilisation**

Les technologies d'assistance aux mouvements couvrent un éventail d'usages allant de la rééducation fonctionnelle (48% des dispositifs étudiés) à l'assistance aux activités de vie quotidienne (AVQ) (30%), avec parfois un double usage rééducatif et « assistif » (22%). Certains dispositifs, développés initialement pour compenser un déficit moteur dans les AVQ, se révèlent contributifs en rééducation, en stimulant les membres de manière répétée. Par exemple, le dispositif *HandinMind*[6], un gant robotique électrique destiné à améliorer la préhension en vie quotidienne, présente un impact fonctionnel en stimulant la neuroplasticité et en améliorant la coordination et la force musculaire, illustrant comment une utilisation prolongée peut contribuer à la rééducation. On distingue également les dispositifs qui vont apporter une compensation partielle ou complète, de ceux qui vont se substituer totalement à la fonction. Les dispositifs de compensation visent à améliorer les capacités fonctionnelles résiduelles en réduisant les efforts nécessaires pour effectuer certains mouvements. Par exemple, le *Yumen Arm*[7] un support de bras mécanique soutient l'abduction et la flexion du bras, s'inscrivant dans des usages de compensation pour les patients atteints d'hémiparésie. Les dispositifs de substitution s'adressent aux patients ayant une perte totale de fonction motrice et permettent de remplacer la fonction du MS, la tâche est déléguée à un effecteur. Ces technologies, comme le bras robotique Kinova JACO®[8] permettent aux utilisateurs d'interagir avec leur environnement.

**Segments du MS et mouvements impliqués**

Le mouvement du MS se divise en deux phases principales. La phase d'approche, orchestrée par l'épaule et le coude, positionne le MS dans l'espace pour permettre l'interaction avec une cible ou un objet. Ensuite, la phase de saisie mobilise le poignet et la main pour réaliser la préhension et la manipulation. Ces mouvements s'appuient sur une boucle sensori-motrice essentielle, où les retours sensoriels tels que la proprioception et le toucher ajustent en temps réel la coordination. Toute perturbation dans cette interaction entre les systèmes sensoriels, moteurs et cognitifs peut affecter l'exécution des gestes fonctionnels, soulignant l'importance d'une coordination fluide et globale. Les dispositifs d'assistance vont chercher à répondre aux besoins fonctionnels des patients dans leur complexité, en s'adressant à la facilitation de la phase d'approche, de la phase de saisie, ou de l'ensemble.

Selon la littérature, le segment distal (main, doigts, poignet), indispensable à la préhension, est la cible principale (46 %). Par exemple, le gant robotisé *J-Glove*[9] aide à restaurer les fonctions de saisie, de relâchement ou de manipulation d'objets. Une approche proximo-distale, adoptée par 24 % des dispositifs, vise à coordonner les segments proximaux (épaule, coude) et distaux pour des mouvements fluides, souvent à travers des systèmes de compensation gravitationnelle. Le dispositif *SpringWear*[10] illustre cela en soutenant mécaniquement l'épaule et le coude, pour réduire l'effort proximal et faciliter les gestes fins. Les dispositifs couvrant le MS dans son intégralité (de la main à l'épaule) sont représentés à 27%, avec, par exemple, le *RUPERT*[11], qui soutient les articulations de l'épaule aux doigts. Les solutions limitées aux segments proximaux, comme le coude, sont plus rares dans la littérature. Un exemple est le *HAL-SJ*[12], un exosquelette motorisé qui facilite la flexion et l'extension du coude.

**Technologies employées**

L'analyse des dispositifs d'assistance aux mouvements pour le MS révèle une grande diversité de conceptions et d'applications. Les terminologies et leurs classifications peuvent parfois être sujettes à interprétation. Certains auteurs assimilent les supports de bras à des exosquelettes, et inversement. La frontière avec les orthèses est également fragile. Voici les définitions que nous utiliserons :

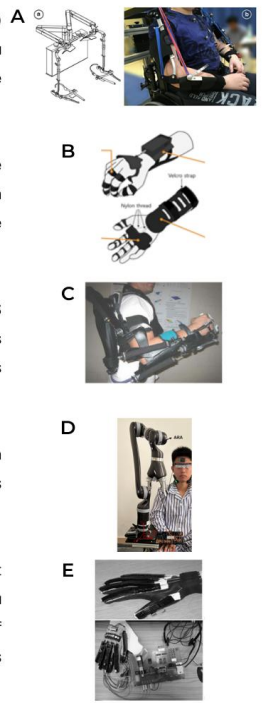

- **Support de bras** : dispositif conçu pour réduire les efforts nécessaires à la mobilisation du membre supérieur (MS) en compensant la gravité. Contrairement aux solutions ciblant des segments spécifiques, il soutient l'ensemble du bras pour faciliter son mouvement. Son interface avec le MS se présente généralement sous la forme d'une manchette antébrachiale, assurant un contact léger et homogène avec le bras.

- **Orthèse** : dispositif spécifiquement destiné à corriger, stabiliser ou améliorer une fonction motrice déficiente. Elle agit directement sur une articulation ou un segment précis du MS, comme le poignet, le coude ou la main. Son interface est souvent constituée d'une coque rigide ou semi-rigide, ou encore de sangles ajustées, permettant une fixation localisée et précise pour un maintien optimal.

- **Exosquelette** : dispositif mécanisé complexe, souvent motorisé, conçu pour assister plusieurs articulations du MS (épaule, coude, poignet, main). Il s'applique directement sur le bras avec des points d'appuis sur les différents segments, permettant de transmettre les efforts mécaniques nécessaires pour amplifier, guider ou reproduire des mouvements.

- **Robot manipulateur** : dispositif mécatronique capable de réaliser des tâches pré-définies, automatiquement ou en mode télé-programmé ou télé-opéré, pour la saisie, la préhension et la manipulation d'objet. Diverses interfaces permettent un pilotage direct ou semi automatisé.

- **Gant robotique** : dispositif conçu pour soutenir, restaurer ou améliorer les fonctions motrices de la main. Il s'agit généralement d'une structure souple ou semi-rigide avec des actionneurs et des capteurs, permettant d'assister ou de reproduire les mouvements des doigts. L'interface d'un gant robotique avec le MS est directe : le dispositif enveloppe la main et les doigts, et peut inclure des éléments sur l'avant-bras pour ancrer le mécanisme ou loger les composants électroniques.

Figure [2] : Définitions et classification des technologies étudiées dans cet article avec pour illustration (A) Support de bras mécanique Yumen Arm ; (B) Orthèse électrique ; (C) Exosquelette électrique RUPERT ; (D) Bras Robotique JACO ; (E) Gant robotique J-Glove

Parmi ces dispositifs, on distinguera également les dispositifs mécaniques, qui utilisent des ressorts, leviers ou câbles pour compenser la gravité et réduire l'effort musculaire, des dispositifs électriques, qui s'appuient sur des moteurs, capteurs ou systèmes électroniques avancés pour offrir une assistance plus active et programmable.

Les dispositifs mécaniques représentent 33% des dispositifs étudiés. On retrouve :
1. Les supports de bras (18%) tels que le *Yumen Arm*$_{(7)}$ qui offre un soutien à l'abduction et la flexion du bras.
2. Les exosquelettes (11%) comme le *HandSOME II*$_{(13)}$ utilisant des ressorts élastiques pour faciliter l'ouverture des doigts après une préhension.
3. Les gants robotiques (4%) tels que le *SaeboGlove*$_{(14)}$, qui utilise un système de ressorts extensibles pour faciliter l'extension des doigts, aidant à la préhension, au relâchement et à la manipulation d'objets.

Les dispositifs électriques quant à eux représentent 67 % des dispositifs étudiés, on distingue :
1. Les exosquelettes (26%) tels que le *RUPERT*$_{(11)}$, qui offrent une assistance ciblée et programmable grâce à des systèmes électroniques avancés.
2. Les gants robotiques (22%) ciblant les fonctions de la main. Le gant *J-Glove*$_{(9)}$, par exemple, utilise les signaux électromyographiques (EMG) pour déclencher l'ouverture et la fermeture de la main.
3. Les robots (11%) tels que le bras robotique Kinova JACO®$_{(8)}$, intégrant une fonction de préhension pour permettre des mouvements précis et la manipulation d'objets. Les membres surnuméraires (*Soft-SixthFinger*$_{(15)}$ et *Soft Hand X*$_{(16)}$) ont été rajoutés dans cette catégorie.
4. Les supports de bras électriques (4%) sont illustrés par le *HapticMaster*$_{(17)}$, qui peut assister l'utilisateur dans des trajectoires prédéfinies tout en ajustant l'intensité des forces appliquées.
5. Les orthèses électriques (4%) sont représentées uniquement par *Yoo et al*$_{(18)}$, qui fait usage d'une orthèse dynamique basée sur le phénomène de ténodèse.

Selon les besoins, différents types d'actionneurs sont intégrés aux dispositifs électriques :
1. Actionneurs hydrauliques : adaptés pour générer de grandes forces avec précision, ces systèmes restent encombrants et nécessitent un entretien rigoureux, les rendant moins adaptés aux dispositifs portables. Aucun article de cette revue n'en a rapporté l'usage.
2. Actionneurs pneumatiques : utilisés dans des systèmes tels que le *RUPERT*$_{(11)}$, ils offrent une bonne souplesse et rapidité de réponse, mais nécessitent des compresseurs ou réservoirs, limitant leur portabilité.
3. Actionneurs électriques : fréquemment adoptés, ils allient polyvalence et précision, comme illustré avec le *Springwear*$_{(10)}$. Cependant, la force produite reste limitée par les capacités des batteries, rendant leur autonomie perfectible.

**Commandes et interfaces de pilotage**
Les dispositifs d'assistance pour le MS proposent divers systèmes de commandes et interfaces, adaptés aux capacités des utilisateurs et aux usages envisagés. Deux grandes catégories se distinguent :
- Les commandes directes, nécessitant une action explicite de l'utilisateur, comme les commandes mécaniques (45 %), qui incluent des leviers, boutons ou interrupteurs ainsi que des systèmes de force ou de pression. Ces solutions, souvent intégrées dans des dispositifs simples comme le *Yumen Arm*$_{(7)}$, sont appréciées pour leur robustesse et leur accessibilité. Les interfaces tactiles (7 %) favorisent une interaction par contact ou commande vocale. Enfin, 7 % des dispositifs combinent commandes mécaniques et électroniques dans des systèmes manuels assistés.

- Les commandes par détection d'intention exploitent des signaux biomécaniques ou neurophysiologiques pour contrôler le dispositif. Les commandes biologiques (30%), comme dans l'exosquelette *HAL-SJ*$_{(12)}$, exploitent des signaux musculaires résiduels. Certaines interfaces hybrides combinent plusieurs technologies, comme dans l'article de *Lin et al*, qui associe le bras robotique Kinova JACO®$_{(8)}$ avec une interface de contrôle intégrant le suivi du regard et l'EMG facial via une architecture réseau pour le « deep learning » appelée réseau neuronal convolutif. Les capteurs de position (7 %) ajustent aussi les mouvements en temps réel.

**Pathologies adressées**

Les dispositifs d'assistance aux mouvements pour le MS ciblent diverses pathologies, avec une représentation importante de l'AVC à la phase chronique (56%), une cause majeure de déficience motrice. Parmi les solutions dédiées à ce groupe, on retrouve l'exosquelette électrique *TIGER*$_{(19)}$, ou le gant *SaeboGlove*$_{(14)}$. Pour les patients en phase aiguë post-AVC (11 % des cas), les dispositifs tels que le gant robotique *HandinMind*$_{(6)}$ favorisent une rééducation intensive, stimulant la plasticité cérébrale et renforçant les connexions neuromusculaires. Les lésions de la moelle épinière représentent 22 % des cas, souvent caractérisées par une paralysie partielle ou totale. Les bras robotiques et gants souples robotisés y jouent un rôle clé, offrant une compensation fonctionnelle partielle ou complète. Les pathologies neuromusculaires (7%), telles que la dystrophie musculaire ou l'amyotrophie spinale, sont marquées par une perte progressive de la force musculaire, nécessitant des solutions compensatoires comme le *Yumen Arm*$_{(7)}$ ou le *SpringWear*$_{(10)}$, qui aident à maintenir une certaine autonomie dans les tâches quotidiennes. Enfin, les patients atteints d'autres troubles neurologiques (4%), tels que la paralysie cérébrale, bénéficient de dispositifs comme le WREX®(20) conçu pour gérer les contractions spastiques et faciliter des mouvements plus fluides.

**Avantages et limites**

Les dispositifs d'assistance aux mouvements offrent divers avantages qui mettent en avant leur efficacité clinique et leur intérêt croissant. Leur premier atout réside dans l'amélioration des capacités fonctionnelles dans les AVQ. Par exemple, le Kinova JACO®$_{(8)}$ offre une substitution fonctionnelle permettant d'effectuer des tâches complexes. Des supports gravitaires, tel que le WREX®$_{(20)}$, réduisent l'effort musculaire, un avantage notable pour les patients post-AVC ou atteints de pathologies neuromusculaires confrontés à la fatigue musculaires. L'adaptabilité est un autre atout essentiel : des solutions comme le *J-Glove*$_{(9)}$ ou le *SaeboGlove*$_{(14)}$ s'ajustent aux besoins spécifiques, élargissant leur pertinence en milieu clinique et à domicile. Enfin, certaines technologies, telles que le *TIGER*$_{(19)}$ ou le *HAL-SJ*$_{(12)}$ associent assistance intelligente et portabilité, répondant aux besoins variés des utilisateurs.

Cependant, ces technologies présentent des limites notables, qui entravent parfois leur adoption. La dépendance à l'activité musculaire résiduelle rend certains dispositifs, comme le *J-Glove*$_{(9)}$, inadaptés aux paralysies complètes. Les contraintes ergonomiques sont également perçues : certains dispositifs, tels que le WREX®(20) ou le *TIGER*$_{(19)}$, peuvent être encombrants ou inconfortables. De plus, les solutions standardisées peinent à répondre à la diversité des pathologies. Les dispositifs passifs, bien qu'accessibles, ne répondent pas aux besoins des mouvements complexes ou des déficits sévères. Certaines solutions comme les gants robotiques, ciblent exclusivement la main sans assistance pour les articulations proximales. Enfin, les coûts élevés de dispositifs avancés, tels que le Kinova JACO®$_{(8)}$, et leur dépendance énergétique limitent leur acceptabilité et leur usage en vie quotidienne.

**Discussion**

Malgré les avancées réalisées dans le domaine des dispositifs d'assistance pour le MS, plusieurs limitations demeurent. Une tendance notable est la prédominance des dispositifs motorisés, comme les bras robotiques et les gants motorisés, qui offrent une compensation active et puissante des déficiences motrices. Cependant, cette approche soulève des défis importants : ces dispositifs sont souvent coûteux, énergivores et leur adoption reste limitée, notamment en dehors des environnements cliniques, en raison de leur complexité d'utilisation.

Un autre constat majeur de cette revue est le manque de littérature sur les dispositifs actuellement commercialisés. Bien que quelques études aient été publiées sur des technologies comme le bras robotique JACO®, l'évaluation des dispositifs en contextes réels demeure insuffisante. Cela représente une réelle opportunité de recherche, permettant non seulement de mieux comprendre les bénéfices concrets pour les patients, mais aussi d'identifier les limites de ces technologies et d'encourager des démarches médico-économiques pour favoriser leur remboursement. Cette recherche pourrait améliorer l'adoption des aides techniques et étendre leur accessibilité à une plus grande population.

De plus, la littérature se concentre souvent sur une seule pathologie, négligeant la diversité des situations qui pourraient être adressées par ces dispositifs. Les pathologies évolutives exigent des solutions modulaires et polyvalentes capables de s'adapter aux besoins changeants des utilisateurs. En ce sens, une approche modulaire pourrait maximiser l'utilité des dispositifs et encourager leur adoption.

Enfin, il devient crucial d'orienter les développements futurs en renforçant les itérations entre usagers, cliniciens et industriels. Ces co-conceptions pourraient garantir que les besoins des utilisateurs soient intégrés dès les premières phases de conception, augmentant ainsi les chances d'aboutir à des dispositifs viables, utiles, et accessibles.

**Conclusion et perspectives**

Cette revue de littérature a exploré l'état actuel des dispositifs d'assistance pour le MS, mettant en lumière leur potentiel pour répondre aux besoins des personnes en situation de handicap. Cependant, leur adoption et efficacité restent limitées par des verrous technologiques et structurels. Des perspectives prometteuses émergent, comme l'utilisation de matériaux avancés, tels que des textiles légers avec capteurs et actionneurs, pour améliorer le confort et la portabilité des dispositifs. Les interfaces multimodales combinant signaux myoélectriques, reconnaissance de mouvements et intelligence artificielle permettraient des systèmes de commande plus intuitifs. L'intégration de technologies éco-énergétiques prolongerait l'autonomie des dispositifs tout en réduisant leur impact environnemental. Le développement de solutions assistant plusieurs articulations offrirait des dispositifs plus polyvalents, répondant à divers besoins. Enfin, la co-conception entre utilisateurs, industriels et professionnels est essentielle pour garantir la pertinence et l'adoption de ces technologies, visant à atteindre un niveau de maturité technologique plus élevé et en assurer la diffusion. En somme, l'avenir des dispositifs d'assistance réside dans des technologies accessibles, durables, polyvalentes et centrées sur les usagers, favorisant une réadaptation fonctionnelle et sociale pérenne.